\documentclass{article}
\usepackage{spconf,amsmath,graphicx}

\usepackage[htt]{hyphenat}
\usepackage{amsmath,amsfonts,bm}

\def\eqref#1{equation~\ref{#1}}
\def\1{\bm{1}}

\def\vx{{\bm{x}}}

\def\mA{{\bm{A}}}

\def\mS{{\bm{S}}}

\def\mW{{\bm{W}}}
\def\mX{{\bm{X}}}
\def\mY{{\bm{Y}}}

\DeclareMathAlphabet{\mathsfit}{\encodingdefault}{\sfdefault}{m}{sl}
\SetMathAlphabet{\mathsfit}{bold}{\encodingdefault}{\sfdefault}{bx}{n}
\newcommand{\tens}[1]{\bm{\mathsfit{#1}}}

\def\tX{{\tens{X}}}

\newcommand{\R}{\mathbb{R}}

\DeclareMathOperator*{\argmin}{arg\,min}

\usepackage{multicol}
\usepackage{multirow}
\usepackage{makecell}
\usepackage{hyperref}
\usepackage[ruled]{algorithm}
\usepackage[noend]{algpseudocode}

\usepackage{xcolor}[dvipsnames]
\usepackage{kpfonts}

\newcommand{\edit}[1]{{#1}}

\definecolor{joshcolor}{rgb}{0.25, 0.75, 0.55}
  
\newcommand{\cL}{{\mathcal{L}}}

\title{A Generalized Hierarchical Nonnegative Tensor Decomposition}
\twoauthors
  {\hspace{3cm}Joshua Vendrow$^\star$, Jamie Haddock$^\dagger$, Deanna Needell$^\star$\thanks{The authors were partially supported by NSF DMS $\#2011140$ and BIGDATA $\#1740325$. JH is also partially supported by DMS \#2211318.}}
	{\hspace{-4cm}$^\star$University of California, Los Angeles\\ \hspace{-4cm}Department of Mathematics\\ \hspace{-4cm}520 Portola Plaza, Los Angeles, CA 90095}
  {}
	{\hspace{-5.5cm}$^\dagger$Harvey Mudd College\\
	\hspace{-5.5cm}Department of Mathematics\\
	\hspace{-5.5cm}301 Platt Blvd, Claremont, CA 91711}
\begin{document}
%\ninept
%
\maketitle

\begin{abstract}
Nonnegative matrix factorization (NMF) has found many applications including topic modeling and document analysis. Hierarchical NMF (HNMF) variants are able to learn topics at various levels of granularity and illustrate their hierarchical relationship. Recently, nonnegative tensor factorization (NTF) methods have been applied in a similar fashion in order to handle data sets with complex, multi-modal structure. Hierarchical NTF (HNTF) methods have been proposed, however these methods do not naturally generalize their matrix-based counterparts. Here, we propose a new HNTF model which directly generalizes a HNMF model special case, and provide a supervised extension. We also provide a multiplicative updates training method for this model. Our experimental results show that this model more naturally illuminates the topic hierarchy than previous HNMF and HNTF methods.
\end{abstract}
\begin{figure*}
    \centering
    \includegraphics[width=0.9\linewidth]{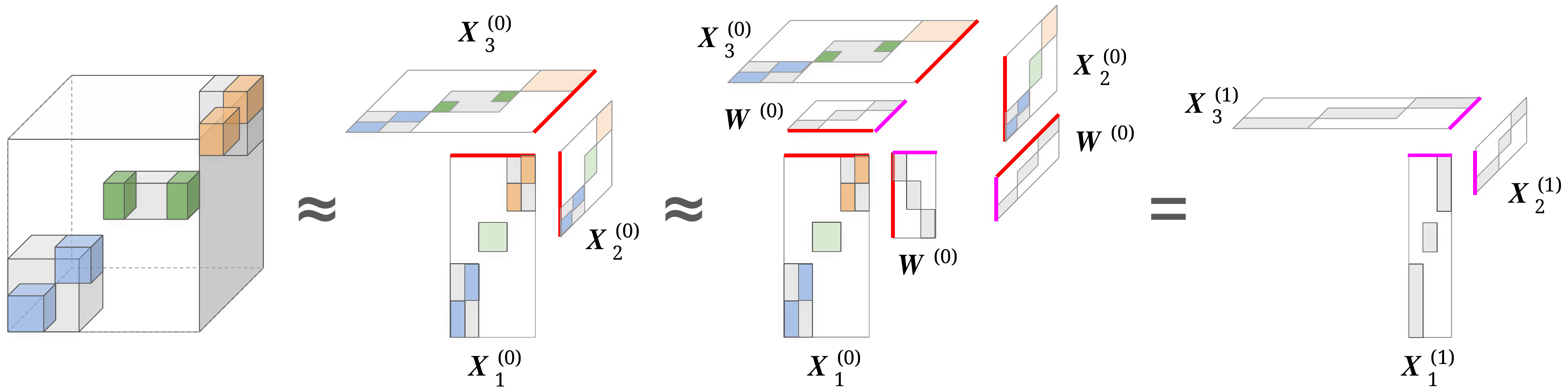}
    \caption{Visualization of a two layer Multi-HNTF on a tensor with three modes.}
    \label{fig:visual}
\end{figure*}
\section{Introduction} \label{sec:intro}
The complexity and size of available data continues to grow which in turn leads to an increasing demand for methods to \edit{interpret} these large data sets. One important task is \textit{topic modelling}, whose goal is to identify latent topics or trends within a set of data. One popular topic modeling approach is nonnegative matrix factorization (NMF), a dimensionality reduction technique that has had great success in the areas of document analysis, clustering, and classification~\cite{lee1999learning, KCP15}. NMF is generalized to multi-modal data by the Nonnegative \edit{CANDECOMP/PARAFAC (CP)} Decomposition (NCPD)~\cite{carroll1970analysis, harshman1970foundations}.

In topic modeling, one often wishes to additionally identify hierarchical relationships between topics learned at different granularities. 
%In order to identify topics at various levels of granularity while identifying relationships between topics
%learned from each separate factorization at high granularity (subtopics) and low granularity (supertopics),
%It is possible to pick up topics at different granularities by applying NMF with different ranks $r$. However, within a forced or encouraged relationship between these factorization, it may be difficult to properly relate the topics learned from each separate factorization at high granularity (subtopics) and low granularity (supertopics). To account for this, 
Towards this goal, many hierarchical models have been developed that enforce an approximate linear relationship between subtopics and supertopics (topics collecting multiple subtopics). More specifically, Hierarchical NMF (HNMF) and Hierarchical \edit{nonnegative tensor factorization} (HNTF) methods have been developed to factorize data sets simultaneously at multiple different granularities with factorizations learned at coarser granularities constrained by the factorizations learned at finer granularities; these sequential factorizations are often viewed as different layers of these models~\cite{FH18,trigeorgis2016deep,GHMNSWZ19,vasilescu2019compositional,song2013hierarchical,grasedyck2010hierarchical,asil21}. 

The most popular models for HNMF and HNTF share a few common issues: (1) these models only use partial information from the factorization at the previous layer in the subsequent layer's factorization; (2) the hierarchical relationships learned often only represent the latent structure in a subset of the modes of the data; and (3) there is no unified model for both matrices and tensors. In order to address these concerns, we develop Multi-HNTF, which provides a unified model for both matrices and tensors, and demonstrates significant improvement in learning latent hierarchical structures in multi-modal data.  

\subsection{Notation.} 
We follow the notational conventions of \cite{goodfellow2016deep}; e.g., tensor $\tX$, matrix $\mX$, vector $\vx$, and (integer or real) scalar $x$.  %In all models, we use variable $r$ (with superscripts denoting layer of hierarchical models) to denote model rank and use $j$ when indexing through rank-one components.  In all tensor decomposition models, we use $r$ to denote the order (number of modes) of the tensor and use $i$ when indexing through modes of the tensor.  In all hierarchical models, we use $\cL$ to denote the number of layers in the model and use $\ell$ to index layers.  
We let $\otimes$ denote the vector outer product and adopt the CP decomposition notation
\begin{equation}[\![\mX_1,\mX_2, \cdots, \mX_k]\!] \equiv \sum \limits_{ j= 1}^{r} \vx_j^{(1)} \otimes \vx_j^{(2)} \otimes \cdots \otimes \vx_j^{(k)},\end{equation}
where $\vx_j^{(i)}$ is the $j$th column of the $i$th factor matrix $\mX_i$ \cite{kolda2009tensor}. %\jh{Xtilde}
%\commjv{We either need to adjust the notation here or adapt this notation throughout the paper}

\subsection{Nonnegative matrix factorization}
NMF is an approach typically applied in unsupervised tasks such as dimensionality-reduction, latent topic modeling, and clustering.  Given nonnegative data matrix $\mX \in \R_{\ge 0}^{m \times n}$ and a user-defined target dimension $r \in \mathbb{N}$ with $r < \min\{m,n\}$, NMF seeks nonnegative factor matrices $\mA \in \R_{\ge 0}^{m \times r}$, often referred to as the \emph{dictionary} or \emph{topic} matrix, and $\mS \in \R_{\ge 0}^{r \times n}$, often referred to as the \emph{representation} or \emph{coefficient matrix}, such that $\mX \approx \mA \mS$.  There are many formulations of this model (see e.g.,~\cite{cichocki2009nonnegative,lee1999learning,lee2001algorithms}) but the most popular utilizes the Frobenius norm, %formulation, 
\begin{equation}\label{eq:fro_NMF}
\argmin_{\mA\geq 0,\mS\geq 0} \|\mX - \mA \mS\|_F^2.
\end{equation}
%One of the most simple and popular approaches to training this non-convex model is the multiplicative update method~\cite{lee2001algorithms}, although there are numerous others (see e.g., \cite{cichocki2009nonnegative,kim2008fast,kim2008nonnegative,lin2007projected}).
%
Here and throughout, $\mA \ge 0$ denotes the constraint that $\mA$ is entry-wise nonnegative.
%The user-defined parameter $r$, which represents the target dimension or the number of believed latent topics, governs the quality of reconstruction of the data; generally $r$ is chosen so that $r < \min\{m,n\}$ to ensure non-triviality of the factorization.  
The columns of $\mA$ are often referred to as \emph{topics}; the NMF approximations to the data (columns of $\mX$) are additive nonnegative combinations of these topic vectors.  This property of NMF approximations yields interpretability since the strength of relationship between a given data point (column of $\mX$) and the topics of $\mA$ is given in the coefficient vector (corresponding column of $\mS$).  For this reason, NMF has found popularity in applications such as document clustering~\cite{xu2003document},  %\cite{gaussier2005relation,shahnaz2006document,xu2003document,berry2005email,pauca2004text}, 
and image and audio processing~\cite{cichocki2006new}, %~\cite{guillamet2002non,hoyer2002non,lee1999learning}, 
%and financial data mining~\cite{de2008analysis}. %, and audio processing~\cite{}.%,  %\cite{cichocki2006new,gemmeke2013exemplar}, 
%and genetics \cite{liu2017regularized}.
Supervised variants of NMF that jointly factorize both the data matrix $\mX$ and a matrix of supervision information (e.g., class labels) $\mY$ have been proposed~\cite{lee2009semi,AGHKKLLMMNSW20}.

%\subsection{Semi-supervised nonnegative matrix factorization}
%SSNMF is a modified variant of NMF that jointly factorizes a data matrix $\mX \in \R_{\ge 0}^{m \times n}$ and a supervision information matrix $\mY \in \R_{\ge 0}^{c \times n}$ with the goal of learning a dimensionality-reduction model and a model for a supervised learning task (e.g., classification).  That is, given data matrix $\mX$, supervision matrix $\mY$, and target dimension $r \in \mathbb{N}$, SSNMF seeks the dictionary matrix $\mA \in \R_{\ge 0}^{m \times r}$, representation matrix $\mS \in \R_{\ge 0}^{r \times n}$, and \emph{supervision matrix} $\mB \in \R_{\ge 0}^{c \times r}$ such that $\mX \approx \mA \mS$ and $\mY \approx \mB \mS$.  The most popular SSNMF formulation~\cite{lee2009semi} employs a weighted combination of Frobenius norm terms, 
%\begin{equation} \label{eq:ssnmf}
%   \argmin\limits_{\mA, \mS, \mB\geq 0} \underbrace{\|\mX - \mA \mS\|_F^2}_\text{Reconstruction Error} + \lambda \underbrace{\|\mY - \mB \mS\|_F^2}_\text{Classification Error};
%\end{equation}
%recently, other formulations have been proposed~\cite{AGHKKLLMMNSW20}.  %These models are trained primarily via multiplicative updates, although other methods could be modified to handle supervision. 

\subsection{Hierarchical nonnegative matrix factorization}

%As of now, the most common bottom-up 
The most popular hierarchical NMF model decomposes a data matrix $\mX$ by repeatedly applying NMF to the $\mS$ matrix output by the previous decomposition, so that each $\mS^{(i)} \approx \mA^{(i+1)}\mS^{(i+1)}$. Given desired ranks $r_0, r_1, ..., r_{\cL}$, this process recursively produces the set of factorizations
\begin{align}
\begin{split}
\mX &\approx \mA^{(0)}\mS^{(0)}, \\ 
\mX &\approx \mA^{(0)}\mA^{(1)}\mS^{(1)}, \\
& \vdots \\
\mX &\approx \mA^{(0)}\mA^{(1)}\cdots \mA^{(\cL)}\mS^{(\cL)}.
\end{split}
\end{align}
%The simplest algorithm for optimizing this model is to begin with an initial factorization and sequentially factorize the $\mS$ matrices with a standard NMF algorithm. 
Many complex optimization algorithms have been proposed to improve the quality of the factorization and minimize cascading errors, and are mostly based upon this model for HNMF \cite{FH18,trigeorgis2016deep,le2015deep,SNT17,GHMNSWZ19}. 
%These methods provide improvements to the reconstruction accuracy and mitigate cascading error throughout the model, however by simultaneously and repeatedly optimizing a large number of matrices, these methods often significantly increase the time complexity for optimization. 

\subsection{Nonnegative CP Decomposition (NCPD).} %Note that NMF specializes matrix factorization to factorizing a nonnegative data matrix into the product of two (lower-dimensional) nonnegative factor matrices.  In the same way, NNCPD specializes the CP decomposition to decomposing a nonnegative data tensor into the sum of rank-one tensors which are the outer product of nonnegative vectors. Nonnegativity is necessary when we desire to preserve  inherent properties of the original tensor data. For example, a tensor of images will have entries representing pixel values that must be nonnegative. 
The NCPD generalizes NMF to higher-order tensors;
specifically, given an order-$k$ tensor $\tX \in \R^{n_1 \times n_2 \times \cdots \times n_k}_{\ge 0}$ and a fixed integer $r$, the approximate NCPD of $\tX$ seeks $\mX_1 \in \mathbb{R}^{n_1 \times r}_{\ge 0}, \mX_2 \in \mathbb{R}^{n_2 \times r}_{\ge 0}, \cdots, \mX_k \in \mathbb{R}^{n_k \times r}_{\ge 0}$ so that
\begin{equation}\label{eq:nncpd}
  \tX \approx [\![\mX_1,\mX_2, \cdots, \mX_k]\!]. %\equiv \sum \limits_{ j= 1}^{r} \vx_j^{(1)} \otimes \vx_j^{(2)} \otimes \cdots \otimes \vx_j^{(k)},
\end{equation}
%where $\vx_j^{(i)}$ denotes the columns of $\mX_i$. 
The $\mX_i$ matrices will be referred to as the NCPD factor matrices. %A nonnegative approximation with fixed $r$ is obtained by approximately minimizing the reconstruction error between $\tX$ and the NCPD reconstruction. %$\hat X =\sum \limits_{  \ell= 1}^{r} a_\ell \otimes b_\ell \otimes c_\ell$ among all the nonnegative vectors. %When the reconstruction error vanishes we say that an exact rank-$r$ NNCPD is obtained. The nonnegative rank, denoted as $\rank_{+}(X)$, is the minimum integer $r*$ so that there exists an exact rank-$r^*$ NNCPD of $X$. In what follows, unless otherwise stated, when we refer to rank of a tensor we are referring to nonnegative rank \edit{and unless indicated otherwise, $\|\cdot\|$ indicates the Frobenius norm}. 
This decomposition has found numerous applications in the area of \emph{dynamic topic modeling} where one seeks to discover topic emergence and evolution~\cite{cichocki2007nonnegative,traore2018non,saha2012learning}.  %Methods for training NMF models can often be generalized to NCPD; for example, multiplicative updates~\cite{welling2001positive} and alternating least-squares~\cite{kim2014algorithms}.

\subsection{Related work}
%Other Hierarchical NMF methods have improved on the performance of a standard HNMF by improving the optimization of the factor matrices and mitigating error propagation through the layers, but all these methods rely on the same underlying model and extra optimization steps increase the time complexity of the model \cite{FH18,trigeorgis2016deep,GHMNSWZ19}. Additionally, p
Previous works have developed hierarchical tensor decomposition models and methods \cite{vasilescu2019compositional,song2013hierarchical,grasedyck2010hierarchical,asil21}. The models most similar to ours are that of \cite{cichocki2007hierarchical}, which we refer to as hierarchical nonnegative tensor factorization (HNTF), and \cite{asil21}, which we refer to as HNCPD. HNTF consists of a sequence of NCPDs, in which one of the factor matrix is held constant at each layer while the remaining factor matrices produce the tensor that is decomposed at the next layer, and as a result the performance of HNTF varies significantly based on which data mode appears first in the representation of the tensor. We refer to `HNTF-$i$` as HNTF applied to the representation of the tensor where the modes are reordered with mode $i$ first. HNCPD consists of an initial NCPD followed by an HNMF applied to each of the resulting factor matrices. In what follows, Neural HNCPD denotes the HNCPD model trained with a neural network architecture, while Standard HNCPD denotes the HNCPD model trained with a multiplicative updates method~\cite{asil21}.%When optimized through a neural network architecture, the resulting Neural HNCPD shows significant improvement over previous methods but has a greater complexity. When optimized using multiplicative updates, the resulting Standard HNCPD has comparable time complexity, but suffers from poor performance because the multiple HNMF branches are unable to work together to optimize the model.  

\subsection{Contribution and Organization}
%\jh{to be updated} \dn{Also please check you like what I wrote here.}
%We offer two contributions. First, we propose a simple model for Hierarchical NMF that offers more flexability than previous models. \jh{Actually, less flexibility?} Second, we demonstrate a natural and direct correspondence between this model and a new model for Hierarchical NTF. 
In Section~\ref{sec:methods}, we introduce our proposed hierarchical tensor decomposition model, Multi-HNTF, first for the special case of matrix data, and then in general for tensor data.  %By introducing a single model that naturally generalizes from matrix to tensor data, we hope to increase the flexibility and simplicity for handling data sets of any shape. We will showcase this benefit by demonstrating experiments on both matrix and tensor data sets. 
%
%Following these objectives, in Section \ref{sec:methods} we propose the our new parallel HNMF and HNTF models. 
In Section \ref{sec:experiments}, we perform topic modeling experiments on two document analysis data sets and one synthetic data set, and compare our model to other HNMF and HNTF models. Finally, in Section \ref{sec:conclusion} we summarize our findings and discuss future work. 

\section{MODEL}
\label{sec:methods}

Here we describe our proposed model for hierarchical matrix and tensor decomposition, Multi-HNTF.  Like previous hierarchical nonnegative matrix factorization models, we seek a series of factorizations in which sequential factor matrices are approximate factors of their predecessors (thus leading to a clear linear relationship between learned topics).  However, unlike previous matrix factorization models, we propose a hierarchical decomposition model that generalizes naturally to higher-order tensors.  As the matrix model is a special case of the tensor decomposition model, we give pseudo-code only for the tensor model but give intuition for the matrix case first.

%\subsection{Heirarchical NMF}

\textbf{Matrix model:} Our model consists of a sequence of nonnegative matrix factorizations that decompose input data matrix $\mX \in \R^{m \times n}$.  Given desired ranks $r_0, r_1, \cdots, r_\mathcal{L}$, this process produces $\mA^{(\ell)} \in \R^{m \times r_\ell}$ and $\mS^{(\ell)} \in \R^{r_\ell \times n}$ such that $\mX \approx \mA^{(\ell)}\mS^{(\ell)}$, while constraining a linear relationship between successive factor matrices, $$\mA^{(\ell+1)} = \mA^{(\ell)}\mW^{(\ell)}, \;\; \mS^{(\ell+1)} = (\mW^{(\ell)})^T\mS^{(\ell)}$$ where $\mW^{(\ell)} \in \R^{r_{\ell} \times r_{\ell+1}}$ for each $\ell = 0 \ldots \cL$.  We note that the matrix $\mW^{(\ell)}$ collects the $r_{\ell}$ subtopics into $r_{\ell+1}$ supertopics at the $\ell$th layer.

%with an initialize rank $r_0$ NMF $\mX \approx \mA^{(0)}\mS^{(0)}$, where $\mA^{(0)} \in \R^{m \times r_0}$ and $\mS^{(0)} \in \R^{r_0 \times n}$. In the following hierarchical layers of factorization, we seek $\mA^{(\ell)} \in \R^{m \times r_\ell}$ and $\mS^{(\ell)} \in \R^{r_\ell \times n}$ such that $\mX \approx \mA^{(\ell)}\mS^{(\ell)}$, while identifying a relationship between these factor matrices and the initial ones. To do so, we introduce matrices $\mW^{(\ell)}, \mV^{(\ell)} \in \R^{r_{\ell} \times r_{\ell+1}}$, where 
%$$\mA^{(\ell+1)} = \mA^{(\ell)}\mW^{(\ell)}, \;\; \mS^{(\ell+1)} = (\mV^{(\ell)})^T\mS^{(\ell)}$$ for each $\ell = 0 \ldots \cL$.
%Now, at each layer $\ell$, we can arrive at good values for $\mA^{(\ell+1)}$ and $\mS^{(\ell+1)}$ by finding $\mW^{(\ell)}, \mV^{(\ell)}$ such that 
%$$\mX \approx \mA^{(\ell)}\mW^{(i)}(\mV^{(\ell)})^T\mS^{(\ell)}.$$
%and then setting $\mA^{(1)}$ and $\mS^{(1)}$ as above. In Algorithm \ref{alg:HNMF} we display the pseudocode for the HNMF process.

%\begin{algorithm}[H]
%	\caption{Hierchical NMF}
%	\label{alg:HNMF}
%	\begin{algorithmic}
%		\Procedure{HNMF}{$\mX$}
%		\State $\mA^{(0)}, \mS^{(0)} \leftarrow \argmin\limits_{A, S \ge 0} \| \mX - \mA\mS \|$
%		\For{$\ell = 0\ldots\cL$} 
%		\State $\mW^{(\ell)}, \mV{(\ell)} \leftarrow \argmin\limits_{W, V \ge 0} \| \mX - \mA^{(\ell)}WV^T\mS^{(\ell)} \|$
%		\State $\mA^{(\ell+1)} = \mA^{(\ell)}\mW^{(\ell)}$
%		\State $\mS^{(\ell+1)} = (\mV^{(\ell)})^T\mS^{(\ell)}$
%		\EndFor
%		\EndProcedure
%	\end{algorithmic}
%\end{algorithm}

%\subsection{Hierarchical NTF}

\textbf{Tensor model:} 
We generalize this hierarchical model to tensor data by applying the same linear relationship between successive factor matrices in a NCPD model (the tensor generalization of NMF).  Given desired ranks $r_0, r_1, \cdots, r_{\cL}$ and input tensor $\tX \in \mathbb{R}^{n_1, n_2, \cdots, n_k}$, this process produces $\mX_1^{(\ell)} \in \mathbb{R}^{n_1 \times r_\ell}, \mX_2^{(\ell)} \in \mathbb{R}^{n_2 \times r_\ell}, \cdots, \mX_k{(\ell)} \in \mathbb{R}^{n_k \times r_\ell}$ such that $\tX \approx [\![\mX^{(\ell)}_1,\mX^{(\ell)}_2, \ldots, \mX^{(\ell)}_k]\!]$. As in the matrix model, we begin with an initial rank $r_0$ decomposition; in this case, it is an initial rank $r_0$ NCPD $\tX \approx [\![\mX^{(0)}_1,\mX^{(0)}_2, \ldots, \mX^{(0)}_k]\!].$
%Based on the formalation of Hierarchical NMF, we can develop a parallel Hierarchical NTF method by replacing the matrices $\mW^{(i)}$ and $\mV^{(i)}$ with a set of matrices $\{\mW^{(\ell)}_i\}_{i=0,\ell=0}^{k,\cL}$. As in HNMF, we begin with an initial rank $r_0$ factorization, $\tX \approx [\![\mX^{(0)}_1,\mX^{(0)}_2, \ldots, \mX^{(0)}_k]\!]$, and wish to find approximations at each rank $r_\ell$ of the form $\tX \approx [\![\mX^{(\ell)}_1,\mX^{(\ell)}_2, \ldots, \mX^{(\ell)}_k]\!]$. As in HNMF, we use our new matrices, $\mW^{(\ell)}_i$, to relate the factor matrices together. In this case, have that 
We then constrain the linear relationship between successive factor matrices, $$\mX^{(\ell+1)}_{i} = \mX^{(\ell)}_{i}\mW^{(\ell)}$$ for each $\ell = 0 \ldots \cL$ and $i = 1 \ldots k$,
%Now, at each layer $\ell$, we can arrive at a good values for each $\mX^{(\ell+1)}_{i}$ by finding $\{\mW^{(\ell)}_{i}\}_{i=1}^k$ such that 
and compute $\mW^{(\ell)}$ such that the new factor matrices $\mX_i^{(\ell+1)}$ form a rank \edit{$r_{\ell+1}$} NCPD,
$$\tX \approx [\![\mX^{(\ell)}_1\mW^{(\ell)},\mX^{(\ell)}_2\mW^{(\ell)}, \cdots, \mX^{(\ell)}_k\mW^{(\ell)}]\!].$$
See Figure~\ref{fig:visual} for a schematic of this model. In Algorithm \ref{alg:HNTF} we display pseudocode for the Multi-HNTF process.  We note that Line~\ref{line:approx} of Algorithm~\ref{alg:HNTF} is only approximate minimization.  One could apply any approximate minimization scheme; we apply a multiplicative updates \cite{lee2001algorithms} and averaging scheme (we omit details here due to space constraints).

\begin{algorithm}[H]
	\caption{Multi-HNTF}
	\label{alg:HNTF}
	\begin{algorithmic}[1]
		\Procedure{Multi-HNTF}{$\tX$}
		%\State $\{\mX^{(0)}_i\}_{i=1}^k \leftarrow \argmin\limits_{\{\mX_i\}_{i=1}^k \ge 0} \| \tX - [\![\mX^{(0)}_1, \ldots, \mX^{(0)}_k]\!] \|$
		\State $\{\mX^{(0)}_i\}_{i=1}^k \leftarrow \text{NCPD}(\tX, r_0)$
		\For{$\ell = 0\ldots\cL$} 
		\State \mbox{$\mW^{(\ell)} \leftarrow \argmin\limits_{\edit{\mW \in \mathbb{R}_+^{r_\ell \times r_{\ell+1}}}} \| \tX-[\![\mX^{(\ell)}_1\mW, \dots, \mX^{(\ell)}_k\mW]\!] \|$}\label{line:approx}
		\For{$i = 0\ldots k$} 
		\State $\mX^{(\ell+1)}_i = \edit{\mX^{(\ell)}_i\mW^{(\ell)}}$
		\EndFor
		\EndFor
		\EndProcedure
	\end{algorithmic}
\end{algorithm}

\section{EXPERIMENTS}
\label{sec:experiments}
Here, we present the results of applying Multi-HNTF to the 20 Newsgroups data set~\cite{KL08}, a synthetic tensor data set~\cite{asil21}, and a Twitter political data set~\cite{DVNPDI7IN_2016}, along with comparisions to Hierarchical NMF, Neural HNCPD, HNCPD, and HNTF. \edit{Our reconstruction loss is the Frobenius norm of difference of the original and reconstructed tensors.} Code can be found in \url{https://github.com/jvendrow/MultiHNTF}.

\begin{figure*}[htb]
    \centering
    \includegraphics[width=0.8\textwidth]{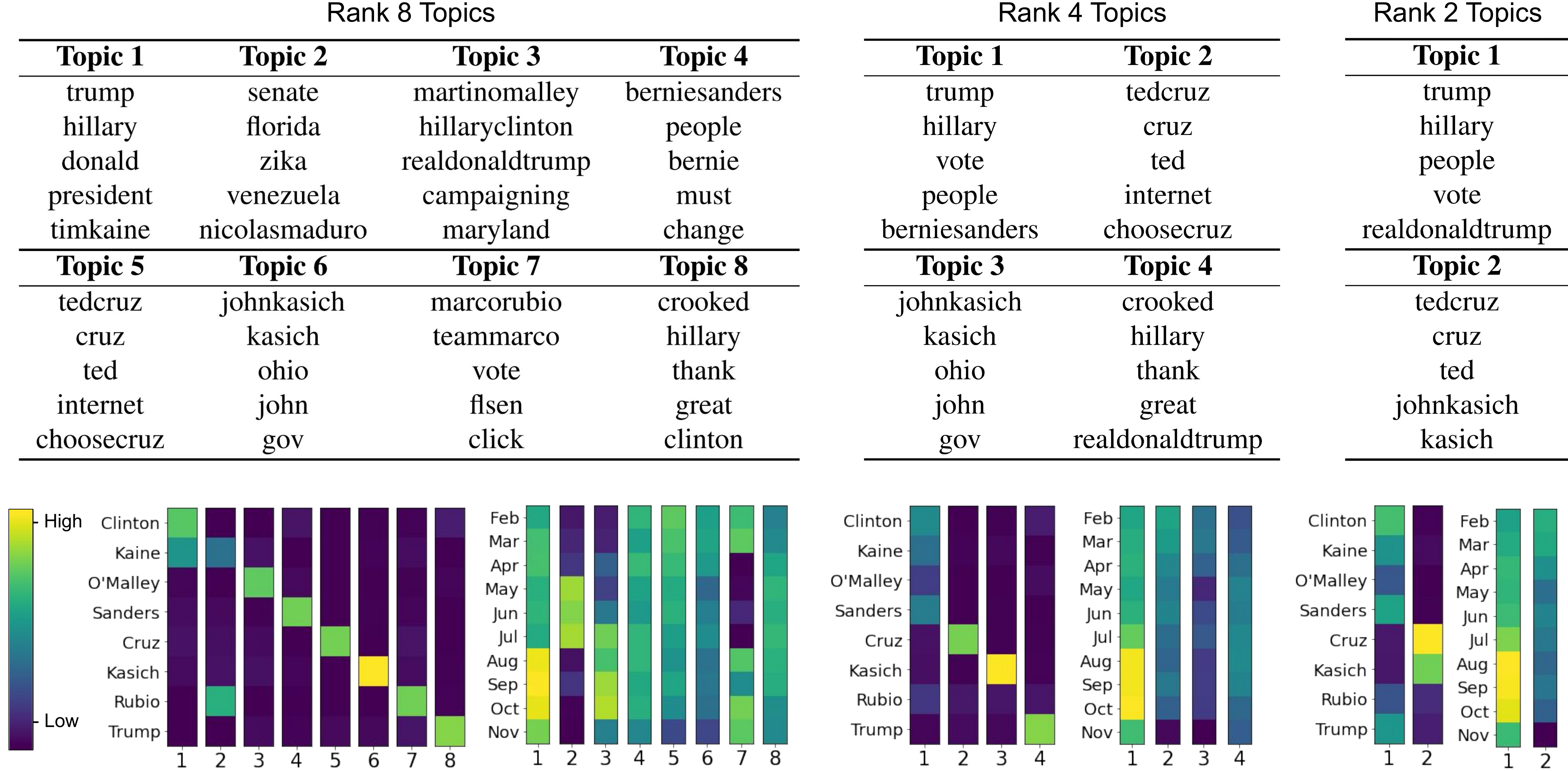}
    \caption{A three-layer Multi-HNTF on the Twitter dataset at ranks $r_0=8$, $r_1=4$ and $r_2=2$.  At each rank, we display the top keywords and topic heatmaps for candidate and temporal modes.}
    \label{fig:twitter}
\end{figure*}

\subsection{20 Newsgroups dataset}

The 20 Newsgroups dataset is a collection of text documents containing messages from newsgroups on the distributed discussion system Usenet~\cite{KL08}. We use a subset of 1000 documents split evenly amongst ten newsgroups (graphics, hardware, forsale, motorcycles, baseball, medicine, space, guns, mideast, and religion) which naturally combine into six supertopics (computer, forsale, recreation, science, politics, religion). We run a two layer Multi-HNTF and HNMF with no supervision, and supervision at both layers, \edit{at ranks $r_0=10$ and $r_1=6$}, and report the results in Table \ref{tab:newsgroups}. We see that with and without supervision, Multi-HNTF outperforms HNMF in reconstruction loss and classification accuracy. %\edit{Need to cut a line here!}

\begin{table}[htb]
	\centering
	\caption{Reconstruction loss and classification accuracy at the second layer of two layer Multi-HNTF and HNMF on the 20 newsgroup data set.}
	\label{tab:20newssub}
	\begin{tabular}{ c | c  c | c  c} 
	    \Xhline{2\arrayrulewidth}
        & \multicolumn{2}{c|}{Recon Loss} & \multicolumn{2}{c}{Accuracy} \\
        Method & Unsup. & Sup. & Unsup. & Sup. \\
        \hline
        Multi-HNTF & $\mathbf{30.81}$ & $\mathbf{30.91}$ & $\mathbf{0.516}$ & $\mathbf{0.737}$\\
        HNMF & $30.82$ & $31.45$ & $0.507$ & $0.636$ \\
		\Xhline{2\arrayrulewidth}
	\end{tabular} 
	\label{tab:newsgroups}
\end{table}

\subsection{Synthetic tensor dataset}
In order to measure the capacity of Multi-HNTF to identify hierarchical relationships \edit{on multi-modal tensor data sets}, we run Multi-HNTF on a synthetic tensor data set introduced in \cite{asil21}. This dataset is a rank seven tensor of size $40 \times 40 \times 40$ comprised of blocks overlayed to form a hierarchical structure, with positive Gaussian noise added to each entry. In Table \ref{tab:synthetic_comparison} we display the relative reconstruction loss on the synthetic dataset for Multi-HNTF and comparable models. We see that Multi-HNTF outperforms Standard HNCPD and every ordering of HNTF at each rank $r_1=4$ and $r_2=2$. Neural HNCPD is able to outperform Multi-HNTF, however due to the repeated forward and backward propagation process, Neural HNCPD utilizes a more complex training method. 
\begin{table}[htb]
    \centering
    \caption{Relative reconstruction loss on the synthetic dataset for Multi-HNTF, Neural HNCPD, Standard HNCPD, and HNTF with two levels of noise over 10 trials. For HNTF we report runs on three re-orderings of the modes the tensor.}%, and for matrix methods we report results for flattening along each mode of the tensor.}
    \begin{tabular}{c|ccc}
        \Xhline{2\arrayrulewidth}
        %\multirow{2}{*}{Method} 
        \multicolumn{1}{c}{Method}
        & $r_0=7$ & $r_1=4$ & $r_2=2$ \\
        \hline
         Multi-HNTF & $0.454$ & $0.548$ & $0.721$ \\  
         Neural HNCPD \cite{asil21} & $0.454$ & $\mathbf{0.508}$ & $\mathbf{0.714}$\\
         Standard HNCPD \cite{asil21} & $0.454$ & $0.612$ & $0.892$\\
         HNTF-$1$ \cite{cichocki2007hierarchical} & $0.454$ & $0.576$ & $0.781$\\
         HNTF-$2$ \cite{cichocki2007hierarchical} & $0.454$ & $0.587$ & $0.765$\\
         HNTF-$3$ \cite{cichocki2007hierarchical} & $0.454$ & $0.560$ & $0.747$\\
         \Xhline{2\arrayrulewidth}

    \end{tabular}
    \label{tab:synthetic_comparison}
\end{table}

\subsection{Twitter political dataset}

The Twitter political data set~\cite{DVNPDI7IN_2016} is a \edit{multi-modal} data set of tweets sent by eight political candidates during the 2016 election season, four Democratic candidates (Hillary Clinton, Tim Kaine, Martin O'Malley, and Bernie Sanders) and four Republican candidates (Ted Cruz, John Kasich, Marco Rubio, and Donald Trump). Following the procedure in \cite{asil21}, we collect all the tweets sent by one politician within each bin of 30 days, from February to December 2016, and combine them into a bag-of-words representation summarizing that politician's twitter activity for the 30 day period. We cap each 30-day bins per politician to 100 tweets to avoid over-fitting to a single month. This forms a tensor of size $8 \times 10 \times 12721$ with 8 politicians, 10 time periods of 30 days, and 12721 words; \edit{multi-modal data with candidate, temporal, and text modes.}

In Table \ref{tab:twitter_comparison}, we list relative reconstruction loss for Multi-HNTF and comparable methods. We see that Multi-HNTF outperforms every method other than Neural HNCPD. In Figure \ref{fig:twitter} we display a visualization of the topics and keywords learned by Multi-HNTF. At rank 8, we see that there is nearly a one-to-one relationship between topics and candidates, at rank 4 many of the democratic candidates combine into a single topic, and at rank 2 Cruz and Kasich are separated from the other candidates. This makes sense because both candidates were Republicans who left the race at similar times. Note that at rank 2, the first topic, which includes the two final presidential candidates, remained strong until the November election, while the topic corresponding to Cruz and Kasich, who dropped out earlier, has weaker presence in later months.

\begin{table}
    \centering
    \caption{Relative reconstruction loss on the Twitter political dataset for Multi-HNTF, Neural HNCPD, Standard NCPD, Standard HNCPD, and 
    HNTF (for each of the possible arrangements of the tensor) at ranks $r_0=8$, $r_1=4$, and $r_2=2$.} %In this experiment we do not we backpropagate to the input for Neural HNCPD, so the approximation at rank 7 are the same for both methods.}
    \begin{tabular}{c|ccc}
        \Xhline{2\arrayrulewidth}
        Method & $r_0=8$ & $r_1=4$ & $r_2=2$ \\
        \hline
         Multi-HNTF & $0.834$ & $0.887$ & $ 0.920$ \\
         Neural HNCPD \cite{asil21} & $0.834$ & $\mathbf{0.883}$ & $\mathbf{0.918}$ \\
         Standard NCPD \cite{asil21} & $0.834$ & $0.889$ & $0.919$\\
         Standard HNCPD & $0.834$ & $0.931$ & $0.950$\\
         HNTF-$1$ \cite{cichocki2007hierarchical} & $0.834$ & $0.890$ & $0.927$  \\
         HNTF-$2$ \cite{cichocki2007hierarchical} & $0.834$ & $0.909$ & $0.956$  \\
         HNTF-$3$ \cite{cichocki2007hierarchical} & $0.834$ & $0.895$ & $0.942$  \\
         \Xhline{2\arrayrulewidth}

    \end{tabular}
    \label{tab:twitter_comparison}
\end{table}

\section{CONCLUSION}
\label{sec:conclusion}
We propose Multi-HNTF, a novel HNTF model that naturally generalizes from a special-case HNMF model. Our initial experiments suggest this model provides improvements both for matrix data over a standard HNMF model, and for \edit{multi-modal} tensor data sets over other HNTF models. We expect that by optimizing our model with a more involved method akin to those of 
\cite{trigeorgis2016deep,le2015deep,GHMNSWZ19,asil21} we could further improve the performance of Multi-HNTF over other models. 

%We propose an NMF-based model, that we call Guided NMF, which incorporates seed topic supervision to guide learned topics towards meaningful and coherent sets of features.  Our initial experiments illustrate the promise of this model in text-based topic modeling applications.  This model could be extended to image/video applications, where the supervision provided encourages object localization and segmentation.
%\dn{This method already rocks the text examples, and likely with some finesse it could be extended to imaging and others.}

\bibliographystyle{IEEEbib}
%\bibliography{strings,refs}
\bibliography{bib}

\end{document}